%% file: paper_arxiv.tex
\documentclass[10pt,twocolumn,letterpaper]{article}

\usepackage{cvpr}              %
\usepackage{graphicx}
\usepackage{amsmath}
\usepackage{amssymb}

\usepackage{scalerel}
\usepackage{placeins}
\usepackage{booktabs}
\usepackage{xcolor}         %
\usepackage{xspace}
\usepackage{enumitem}
\usepackage{adjustbox}
\usepackage[pagebackref,breaklinks,colorlinks]{hyperref}

\usepackage[capitalize]{cleveref}
\crefname{section}{Sec.}{Secs.}
\Crefname{section}{Section}{Sections}
\Crefname{table}{Table}{Tables}
\crefname{table}{Tab.}{Tabs.}

\definecolor{bppgapcol}{RGB}{45,162,170}
\definecolor{red}{RGB}{192,55,44}

\usepackage{soul}

\def\cvprPaperID{6689} %
\def\confName{CVPR}
\def\confYear{2023}

\begin{document}

\title{Multi-Realism Image Compression with a Conditional Generator}%

\author{Eirikur Agustsson\\
Google Research\\
Reykjavík, Iceland \\
{\tt\small eirikur@google.com}
\and
David Minnen\\
Google Research\\
Mountain View, USA \\
{\tt\small dminnen@google.com}
\and
George Toderici\\
Google Research\\
Mountain View, USA \\
{\tt\small gtoderici@google.com}
\and
Fabian Mentzer\\
Google Research\\
Zürich, Switzerland \\
{\tt\small mentzer@google.com}
}
\maketitle

\begin{abstract}
By optimizing the rate-distortion-realism trade-off, generative compression approaches produce detailed, realistic images, even at low bit rates, instead of the blurry reconstructions produced by rate-distortion optimized models. However, previous methods do not explicitly control how much detail is synthesized, which results in a common criticism of these methods: users might be worried that a misleading reconstruction far from the input image is generated. In this work, we alleviate these concerns by training a decoder that can bridge the two regimes and navigate the distortion-realism trade-off. From a single compressed representation, the receiver can decide to either reconstruct a low mean squared error reconstruction that is close to the input, a realistic reconstruction with high perceptual quality, or anything in between. With our method, we set a new state-of-the-art in distortion-realism, pushing the frontier of achievable distortion-realism pairs, i.e., our method achieves better distortions at high realism and better realism at low distortion than ever before.
\end{abstract}

\section{Introduction}
\label{sec:intro}

\begin{figure}[t]
    \centering
    \includegraphics[width=0.69\linewidth]{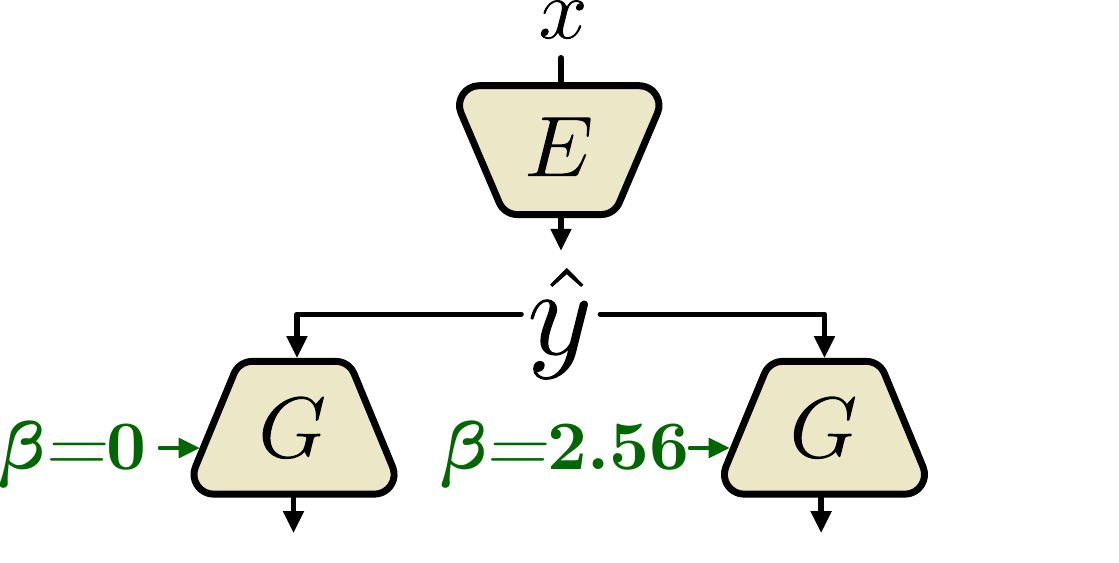} \\[-1ex]
    \begin{tabular}{@{\hskip 0mm}l@{\hskip 0mm}l@{\hskip 0mm}l}
\raisebox{-.5\height}{%
    \includegraphics[width=0.47\linewidth]{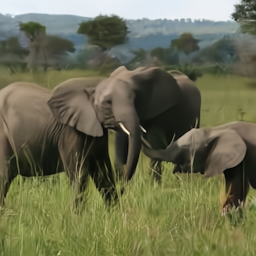}} & \dots & %
\raisebox{-.5\height}{%
\includegraphics[width=0.47\linewidth]{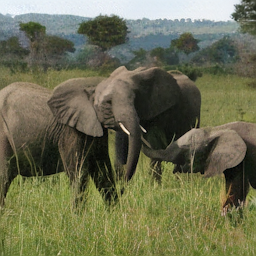}} \\[-3.2ex]
{\color{white} \,28.1dB} 
&&
{\color{white} \,27.0dB}\\[-1ex]

    \end{tabular}
    \caption{\label{fig:teaser}
    Decoding two reconstructions \emph{from the same representation} $\hat y$,
    which takes 2345 bytes to store: We use a low realism weight $\beta=0$ for the left reconstruction, and a high $\beta=2.56$ for the right.
    Note that increasing $\beta$ leads to a much sharper reconstruction, but the PSNR drops by 1.1dB, consistent with rate-distortion-realism theory~\cite{blau2019rethinking}.
   We only show two reconstructions, but our generator $G$ can produce any reconstruction in between
   by changing $\beta$.
    This allows the user to decide between viewing a reconstruction that is close to the input (left, \ie, high PSNR), or that looks realistic (right).
    \vspace{-1.5ex}
    }
\end{figure}

\begin{figure*}
\centering
{
\footnotesize
\begin{tabular}{@{\hskip 0mm}l@{\hskip 1mm}l@{\hskip 0.1mm}l@{\hskip 1mm}l@{\hskip 0mm}}
\toprule
\multicolumn{1}{c}{Input} &
\multicolumn{2}{c}{High-Realism} &
\multicolumn{1}{c}{Low-Distortion} %
\\\midrule

{\scriptsize Kodak: \texttt{kodim20}}
& Ours $\beta{=}2.56$, 0.12bpp, 31.3dB
& HiFiC, 0.12bpp, 29.3dB
& Ours $\beta{=}0$, 0.12bpp, 32.4dB \\
\includegraphics[width=0.245\linewidth]{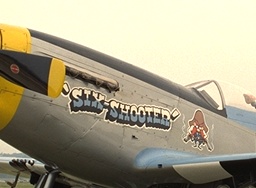} &
\includegraphics[width=0.245\linewidth]{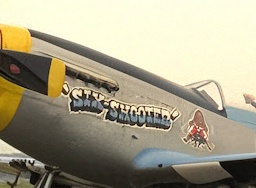} &
\includegraphics[width=0.245\linewidth]{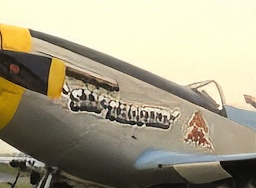} &
\includegraphics[width=0.245\linewidth]{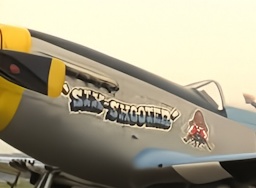} \\
{\scriptsize CLIC 2020: \texttt{3f273}}
& Ours $\beta{=}2.56$, 0.085bpp, 32.6dB
& HiFiC, 0.082bpp, 30.5dB
& Ours $\beta{=}0$, 0.085bpp, 33.6dB \\
\includegraphics[width=0.245\linewidth]{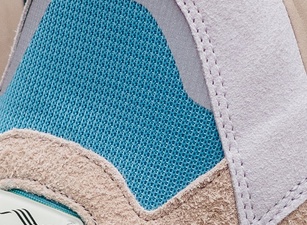} &
\includegraphics[width=0.245\linewidth]{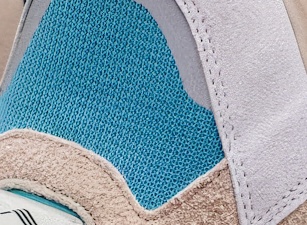} &
\includegraphics[width=0.245\linewidth]{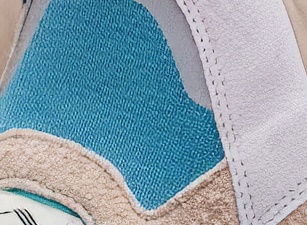} &
\includegraphics[width=0.245\linewidth]{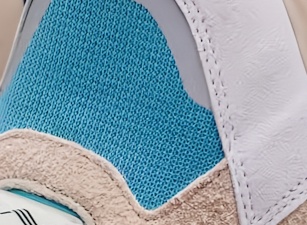} \\
{\scriptsize CLIC 2020: \texttt{88c58}}
& Ours $\beta{=}2.56$, 0.048bpp, 32.3dB
& HiFiC, 0.092bpp {\color{bppgapcol}($\mathbf{1.92{\boldsymbol\times}}$)}, 33.2dB
& Ours $\beta{=}0$, 0.048bpp, 33.7dB \\
\includegraphics[width=0.245\linewidth]{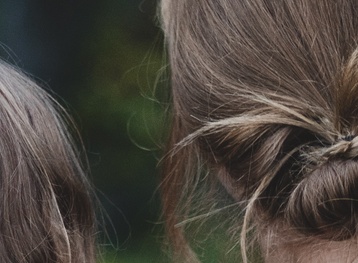} &
\includegraphics[width=0.245\linewidth]{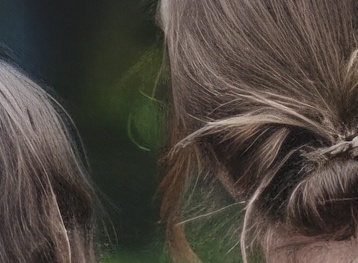} &
\includegraphics[width=0.245\linewidth]{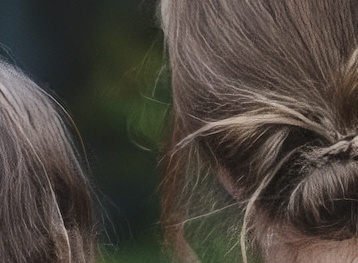} &
\includegraphics[width=0.245\linewidth]{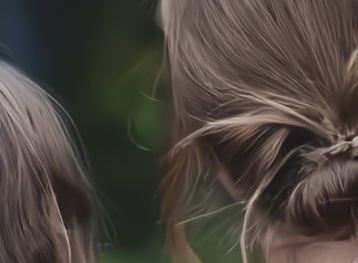} \\
{\scriptsize MS COCO 30K: \texttt{45962}}
& Ours $\beta{=}2.56$, 0.090bpp, 31.0dB
& HiFiC, 0.17bpp {\color{bppgapcol}($\mathbf{1.86{\boldsymbol\times}}$)}, 32.4dB
& Ours $\beta{=}0$, 0.090bpp, 31.9dB \\
\includegraphics[width=0.245\linewidth]{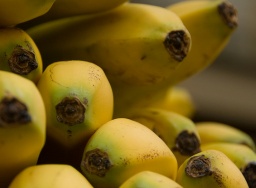} &
\includegraphics[width=0.245\linewidth]{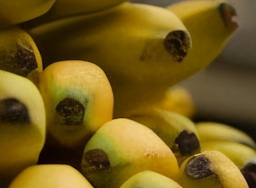} &
\includegraphics[width=0.245\linewidth]{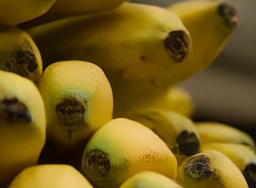} &
\includegraphics[width=0.245\linewidth]{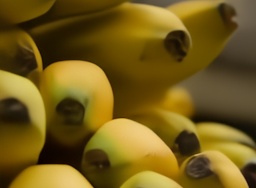} \\[-2ex]

\end{tabular}
}

\caption{\label{fig:visualcomp2}Comparing 
input images to reconstructions from our model at $\beta{=}2.56$, the generative state-of-the-art HiFiC, as well as our model at $\beta{=}0$.
Note that both our models always have the same bits-per-pixel (bpp) per row, since for each row, the two reconstructions we show are obtained from the \emph{same} representation---we simply vary $\beta$ for the generator.
Overall, we see how our high-realism reconstructions ($\beta{=}2.56$) closely match the input, more-so than HiFiC.
On the airplane (first row), we can read the text in our reconstruction, in contrast to the one from HiFiC.
In the second row, the texture of the sneaker is faithfully preserved. 
For the hair, we note that HiFiC uses {\color{bppgapcol}$\mathbf{1.92{\boldsymbol\times}}$} the bitrate of our model to achieve a similar reconstruction. In the last row, HiFiC uses
{\color{bppgapcol}$\mathbf{1.86{\boldsymbol\times}}$} the rate.
In the first two rows, where we have comparble bpp to HiFiC, both of our reconstructions have higher PSNR.
In the rightmost column ($\beta=0$) we can see the low-distortion reconstructions of our model. There we have near state-of-the-art PSNR at the cost of losing the (synthetic) detail.\vspace{-0.5ex}
}
\end{figure*}

Lossy image compression considers the trade-off between the number of bits used to store an input image and how close the reconstruction (that we obtain from the bits) is to that input image.
As we use more bits, we will be able to get closer to the input.
This idea is formalized in the fundamental rate-distortion trade-off~\cite{shannon1959coding}, where ``rate'' stands for bit-rate, and ``distortion'' is formalized as a pair-wise metric between the input image and the reconstruction (\eg, the mean-squared error, MSE).

While minimizing this trade-off has been the focus of many works starting from JPEG~\cite{jpeg1992wallace} all the way to recent neural~\cite{he2022elic} and non-neural~\cite{vtm17} codecs,
there has been a surge of interest in additionally considering the ``realism'' or ``perceptual quality'' of the reconstructions~\cite{blau2019rethinking,tschannen2018deep,theis2021coding,theis2021advantages,mentzer2020high,po-elic,yang2021perceptual,mentzer2021towards,zhang2021universal,theis2022lossy}.
After all, as we move toward low rates, purely rate-\emph{distortion} optimized systems will produce artifacts in the reconstructions, such as the well known block artifacts of JPEG or blurry patches for neural approaches.
There is simply not enough bitrate available to store all of the details, and if we target, \eg, MSE, the best reconstruction is the average image over all images that map to the given representation since, inevitably, many images will map to the same representation at low rates.
Intuitively, instead of an average image reconstruction, we could prefer a ``realistic'' reconstruction that is sharp and appropriately textured. This reconstruction might have worse MSE than the average image, but users might find it more perceptually pleasing and less artificial.
We can see from this argument that there exists an additional trade-off here, between ``realism'' and ``distortion'', and that distortion will increase as we improve realism.
Following Blau and Michaeli~\cite{blau2019rethinking}, we formalize ``distortion'' as a metric between pairs of images (\eg, MSE) that indicates how close is the reconstruction to the input, while
``realism'' indicates how realistic the reconstructions look (regardless of the input).
We formalize the latter as a divergence $d(p_X,p_{\hat X})$ between the distribution of real images, $X$, and reconstructions, $\hat X$. Note that this can only be measured over a set of images since an accurate estimate of the distribution is needed.
Throughout this text, we use PSNR as a measure of distortion, and FID~\cite{heusel2017gans} as a measure of realism.

Previous work successfully optimized the triple rate-distortion-realism trade-off~\cite{agustsson2019extreme,mentzer2020high,po-elic,rippel17a,santurkar2017generative}, however, there is one caveat. Since the realism constraint might produce reconstructions that are far away from the input, these systems might be looked at with suspicion because it is not clear which details
are in the original and which were added by the architecture.

We address this caveat by training a decoder that, given a \emph{single} compressed representation,
either produces a reconstruction where little or no detail is generated (like rate-distortion optimized codecs), one where fine-grained detail is generated (like rate-distortion-realism optimized codecs), or anything in between (see Fig.~\ref{fig:teaser}).
We emphasize that the \emph{receiver} can decide how much detail to generate, because we condition the decoder, not the encoder, on a ``realism factor'', $\beta$, and thus the receiver can produce the full spectrum of reconstructions from a single representation, $\hat y$.

Our main contributions are:
\begin{enumerate}[leftmargin=*,noitemsep]
    \item We bridge the generative and non-generative compression worlds, by navigating the trade-off between distortion and realism from a \emph{single} representation using a conditional generator.
    \item  Our method sets a new state-of-the-art in terms of distortion-realism on high-resolution benchmark datasets,
pushing the frontier
of achievable distortion-realism pairs. 
Our method achieves better distortions  at high realism (low FID) and better realism at low distortion (high PSNR) than ever before (Fig.~\ref{fig:dist_perc}).
\end{enumerate}

\begin{figure*}[t]
    \centering
    \includegraphics[width=\linewidth]{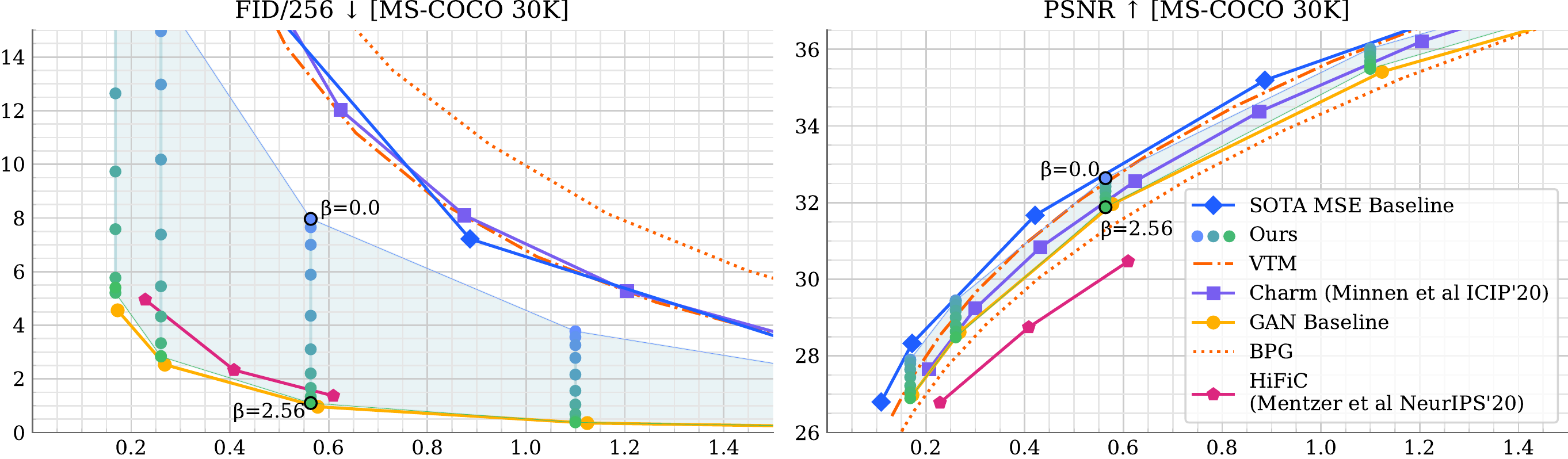}
    \includegraphics[width=\linewidth]{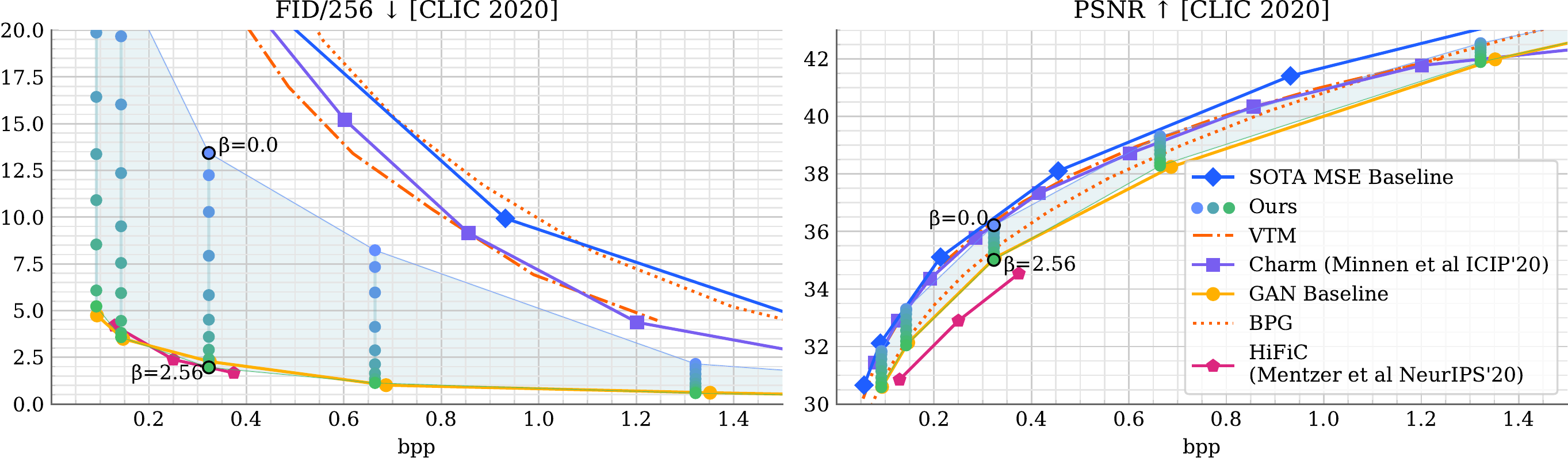}
    \caption{
    \label{fig:results}
    Results on MS-COCO (top) and CLIC 2020 (bottom).
    We show FID as a measure of \emph{realism} (left, lower is better) and PSNR as a measure of \emph{distortion} (right, higher is better).
    The chains of dots 
    \protect\includegraphics[height=2ex]{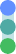}
    indicate the realism-distortion points our method can achieve simply by varying $\beta$ on the receiver side, \ie, we have single model per bitrate.
    We highlight two values of $\beta$ to make this clear.
    We observe that for high $\beta$, we match or outperform HiFiC in FID (realism), while having a significantly better PSNR ($>1$dB).
    On the low $\beta$ side, our model outperforms Charm and matches VTM in terms of PSNR, while having a significantly better FID.
    \vspace{-0.5ex}
    }
\end{figure*}

\section{Related Work}
\label{sec:relwork}

Zhang~\etal~\cite{zhang2021universal} studied ``universal'' representations for rate-distortion-realism and provide a theoretical exploration of the bit-rate overhead incurred by using a \emph{single} representations to obtain different points in the distortion-realism plane.\footnote{Note that the terms ``realism'' and ``perception'' can be used interchangeably. Both refer to a divergence between distributions over images. In contrast, ``perceptual quality'' is used more generally as a measure of subjective visual quality, which could be measured by ``realism'' or by an image quality metric like LPIPS or MS-SSIM.}
They formalize the rate overhead of a universal representation compared to different representations for each distortion-realism point, and show that this overhead is zero for scalar Gaussian sources.
For general sources, they show that the rate-distortion-optimal representation can be used to meet any realism constraint by increasing the distortion by no more than a factor 2. 
They present empirical results on MNIST, showing that after training a single encoder-decoder pair, further decoders can be trained given the frozen encoder. 
We note that we jointly train an encoder and a conditional decoder that navigates the distortion-realism trade-off by adapting to a side input in the form of a single realism factor, $\beta$.

He~\etal present the ``ELIC'' model~\cite{he2022elic}, which is a state-of-the-art neural compression model for rate-distortion (MSE and MS-SSIM~\cite{wang2003multiscale}) performance amongst practical methods. 
Some methods like Koyuncu~\etal~\cite{koyuncu2022contextformer} outperform ELIC at some bit rates, but they use a serial autoregressive context model to improve entropy modeling. Such context models typically lead to 10x slower decode times due to underutilization of the parallel cores of GPUs and TPUs.

Various previous compression methods incorporated adversarial losses to boost realism. Mentzer~\etal developed HiFiC, which combined a conditional GAN~\cite{mirza2014conditional} with a hyperprior-based compression architecture~\cite{balle2018variational} and showed rate savings of 50\% for equal subjective quality compared to MSE-optimized and standard (non-neural) codecs. While ELIC was only optimized for rate-distortion, it was extended to create a ``perceptually-oriented'' variant called PO-ELIC~\cite{po-elic}. This model focused on realism by augmenting the loss function with an adversarial term, a perceptual loss based on LPIPS~\cite{zhang2018unreasonable}, and a patch-based style loss~\cite{gatys2016}. Similarly, Li~\etal~\cite{li2022content} also combine multiple loss terms including a Laplacian loss, MSE, MAE, adversarial loss, and LPIPS, but they merge these terms in a spatially varying, content-adaptive manner based on different detectors (faces, edges, and structure) that run during training. The model is able to learn where to apply each type of loss based on image content, which boosts perceptual quality.
Other methods utilize region-of-interest (ROI) or semantic maps to guide detail and texture synthesis~\cite{Ma2022a,agustsson2019extreme}.

We emphasize that these methods target a single point on the distortion-realism tradeoff, and would require 
storing a different representation for each distortion-realism target.
This is in contrast to our method, which only requires a single model and representation, yet can still generate reconstructions targeting any trade-off along the distortion-realism curve.

More related to our approach, Gao~\etal~\cite{gao2022} present an approach for targeting different multi-distortion trade-offs with a single model using semi-amortized inference: first, a model is trained for a single trade-off to predict a latent representation. This representation is then further optimized for a new trade-off at \emph{encode time}.  Although effective, this approach has several drawbacks:
(1) the new trade-off parameters must be selected at encode time, not decode time and 
(2) encoding becomes very slow since hundreds or thousands of optimization steps must run for each image. %

Iwai~\etal~\cite{iwai2021fidelity} use network interpolation to achieve different distortion-realism targets, however their method operates in a different regime by targeting extremely low bitrates (${<}0.04$bpp on Kodak), where visual quality clearly suffers. We also note that \cite{iwai2021fidelity} is a very different approach: first they train a low-distortion encoder/decoder pair. Then the decoder is finetuned for a GAN loss to synthesize details. At inference time, the model weights of the decoders are interpolated post-hoc.

Theis~\etal~\cite{theis2017lossy} show promising results for generative compression of small (64x64) images using gaussian diffusion and reverse channel coding, obtaining state-of-the-art results on ImageNet64. The approach is based on using reverse channel coding~\cite{theis2022algorithms} to transmit samples, which is under active research and currently computationally prohibitive for the large images we consider here.

\begin{figure}
    \centering
    \includegraphics[width=0.9\linewidth]{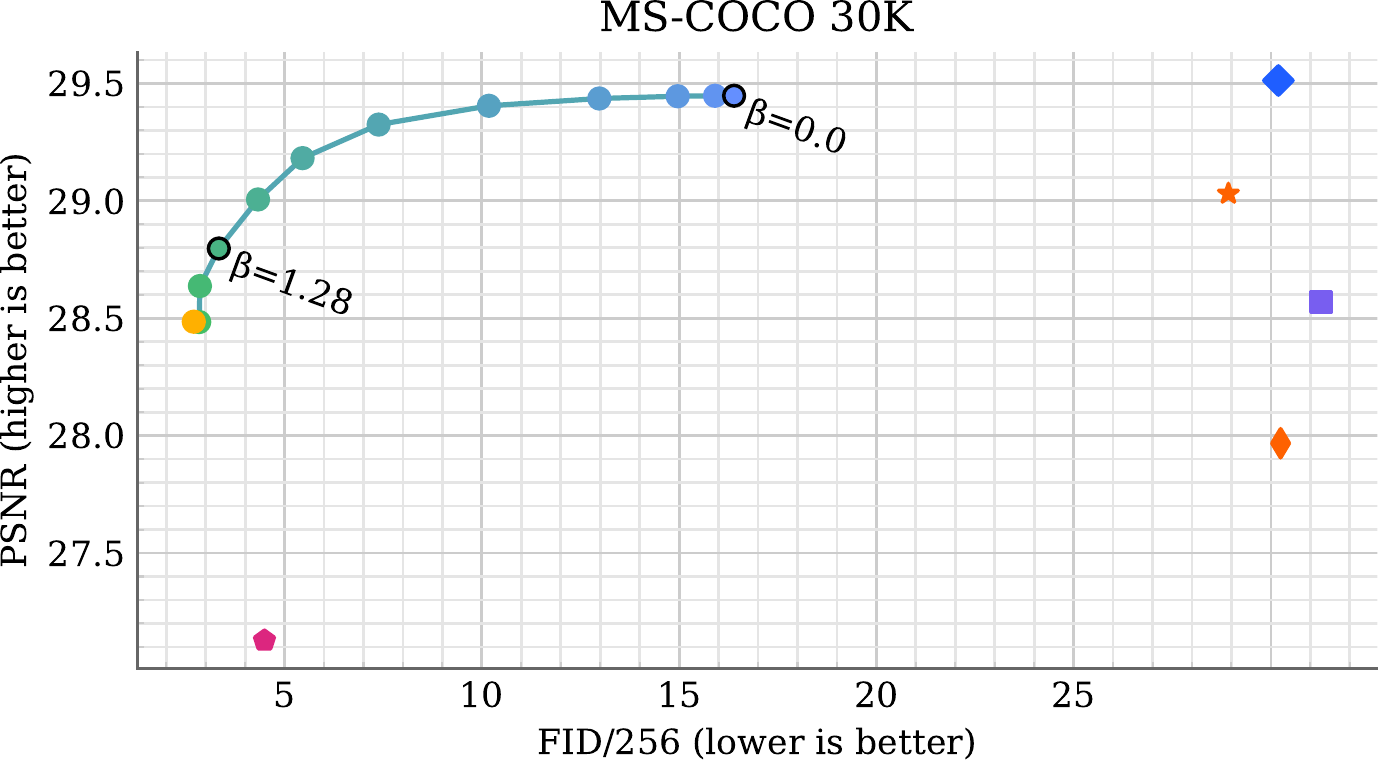}\\[2ex]
    \includegraphics[width=0.9\linewidth]{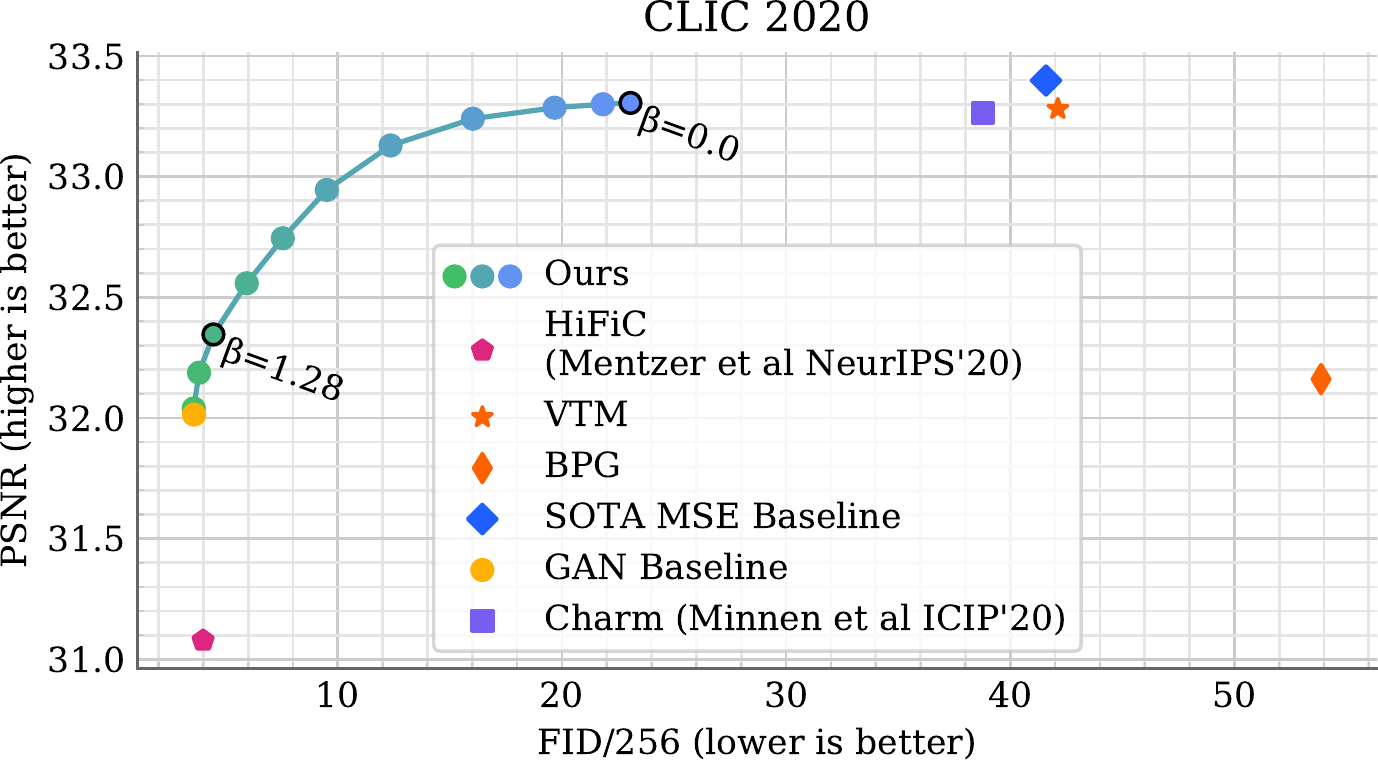}
    \caption{\label{fig:dist_perc}
    Distortion (PSNR) vs.\ realism (FID/256) trade-off on MS-COCO and CLIC 2020, for a \emph{single bitrate}.
    This is obtained by slicing Fig.~\ref{fig:results} at a single bitrate
    (0.26bpp for MS-COCO and 0.14bpp for CLIC 2020).
    In this figure, the optimal point is at the top left, and our method is closer to this optimum than the state-of-the-art generative method HiFiC,
    as well as the SOTA MSE Baseline.
    At high-realism (low FID), we outperform or match HiFiC in FID, while sporting a significantly higher PSNR. 
    At low-distortion (high PSNR), we reach towards state-of-the-art, with significantly lower FID.
    We note that we get a chain of dots for our method because \emph{we can decode multiple reconstructions from the same representation} using different realism weights $\beta$, whereas the shown baselines only have a single dot.
    We see similar results for other bitrates, see Sec~\ref{sec:app:dist_realism_rates}.
    }
\end{figure}

\section{Method}
\label{sec:method}

\subsection{Background}

\paragraph{Neural Image Compression}
We follow the commonly used~\cite{balle2018variational,
minnen2018joint,
mentzer2018conditional,
minnen2020channel,
cheng2020learned,
he2021checkerboard,
he2022elic,
zou2022devil%
} non-linear transform coding approach~\cite{balle2020nonlinear} to do lossy image compression: 
We train an auto-encoder $E,G$, that maps an input image $x$ to a \emph{quantized} representation $\hat y = E(x)$ and back to a reconstruction $\hat x = G(\hat y)$ (we use $G$ for the decoder and call it ``generator'' to avoid confusion with the discriminator $D$ we introduce below).
Training $E,G$ for reconstruction (\eg with a MSE loss) already leads to a compression system, where the sender runs $E$, stores $\hat y$ to disk, the receiver obtains $\hat y$ from disk and runs $G$. However, naively storing $\hat y$ to disk is expensive,
as it takes $\log_2 |\mathcal S|$ bits per symbol (assuming $\hat y_i \in \mathcal S$).
One can do better if the distribution of the symbols is known, as one can then assign shorter bitstrings to more likely symbols.
Given a distribution $p(\hat y)$ estimating the true (unknown) distribution $q(\hat y)$ of the symbols in $\hat y$, we can use entropy coding algorithms to store $\hat y$ using $B(\hat y) = \mathbb{E}_{\hat y \sim q} -\log_2 p(\hat y)$ bits, where $B$ is the cross-entropy between the true data distribution $q$ and our model $p$ (see, \eg, Yang~\etal~\cite{Yang2022a}).
To minimize $B$, recent neural compression approaches train a separate ``entropy model'' $P$ to predict $p(\hat y)$.
Using $P$, we can estimate a bitrate loss during training, $r(\hat y) = B(\hat y)$ and thereby minimize the rate-distortion trade-off by minimizing
\begin{align}
    \mathcal L_{RD} = \mathbb E_{x\sim p_X} [
    r(\hat y) + \lambda \text{MSE}(x, \hat x)], \label{eq:lossmse}
\end{align}
where $\lambda$ controls the trade-off.
Typically, a set of models is trained by varying $\lambda$, which results in models covering different bitrates.

\paragraph{Generative Image Compression}

Inspired by the theoretical formalisation of ``realism'' as a divergence between the distribution of real images $p_X$ and reconstructions $p_{\hat X}$ (see Introduction),
previous works~\cite{agustsson2019extreme,mentzer2020high} use a generative adversarial networks (GANs)~\cite{goodfellow2014generative}-based loss to estimate and minimize this divergence during training.
We follow the formulation of Mentzer~\etal~\cite{mentzer2020high}, which is based on conditional GANs: In addition to $E, G$, a \emph{conditional} discriminator $D(\hat y, x)$ is trained to predict the probability that the given $x$ is a realistic image corresponding to the representation $\hat y$. 
We use the patch discriminator from HiFiC~\cite{mentzer2020high}.
Using $D$, we can formulate the GAN losses for $G$ and $D$ as follows:
\begin{align}
    \mathcal L_G =& \mathbb E_{\hat y \sim p_Y}[-\log ( D(\hat y, G(\hat y)) )] \\
    \mathcal L_D =& \mathbb E_{\hat y \sim p_Y}[-\log (1-D(\hat y,G(\hat y))] + \\ \nonumber
    &\mathbb E_{x\sim p_X }[-\log D(E(x),x)],
\end{align}
where $p_Y$ is the distribution of representations, induced by the encoder transform $E$.

\subsection{Our Approach}\label{sec:ourapproach}

Let $\beta \in [0, \beta_\text{max}]$ be the ``realism weight'' that specifies whether our generator $G$ should produce a low distortion or high realism reconstruction. Our goal is to train a single generator $G$ to work well for any $\beta$. 
To this end, we base our $E,G$ on the ``ELIC'' architectures introduced by He~\etal~\cite{he2022elic} to achieve state-of-the-art rate-distortion results,
but we make $G$ slightly wider, using $N{=}256$ channels instead of $N{=}192$.\footnote{%
        On a one megapixel image using a NVidia V100 GPU, our $G$ runs in 67ms, compared to 43ms for $N{=}192$, and 99ms for HiFiC~\cite{mentzer2020high}.}

\begin{figure}[b]
    \centering
    \includegraphics[width=0.7\linewidth]{%
       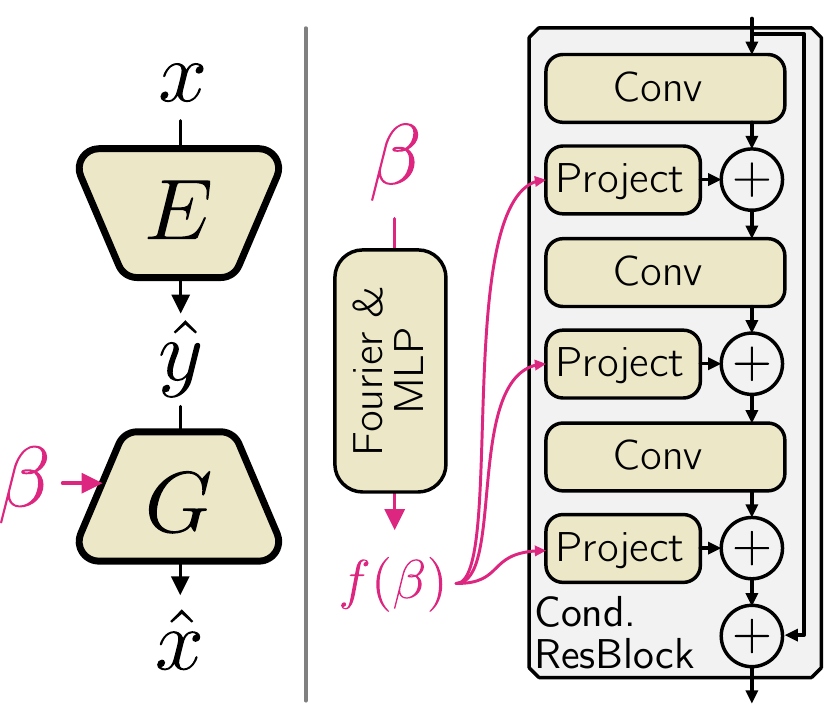}
    \caption{ \label{fig:archoverview} Overview of our architecture. Our encoder $E$ and decoder/generator $G$ are based on the ELIC~\cite{he2022elic} architecture, but we replace every residual block (RB) in $G$ with our conditional  RB shown on the right, to let $G$ know which generative weight $\beta$ to target.
    We first embed $\beta$ using fourier features into a $d$-dimensional space, then apply a 2-layer MLP to obtain features representing $\beta$, $f(\beta) = \text{MLP}(\text{Fourier}(\beta))$.
    Assuming the $i$-th conv.\ layer in the RB has $C_i$ channels, we then project $f(\beta)$ to $C_i$ channels with a learned (per layer) weight $W_i$, obtaining $W_i f(\beta)$. This is added to the output of the conv.\ layer.
       } 
\end{figure}

Additionally, and crucially, we condition $G$ on $\beta$, obtaining a $\beta$-conditional generator $G(\hat y, \beta)$, as shown in Fig.~\ref{fig:archoverview} and described in Sec.~\ref{sec:betacond}. 
Thus, to obtain a reconstruction given a representation $\hat y$, the receiver chooses a $\beta \in [0, \beta_\text{max}]$ and runs $G$ to obtain $\hat x_\beta = G(\hat y, \beta)$.

We adopt the channel autoregressive entropy model (Charm) proposed by Minnen~\etal\cite{minnen2020channel} to minimize bitrate, using 10 slices. 
Combining the losses from the previous section, we obtain the overall loss for training $E,G$ and $D$:
\begin{equation}
 \label{eq:lossbeta}
\begin{split}
    \mathcal L_{EGD}(\beta) = & \mathbb E_{x \sim p_X}\Big[ 
         \lambda' r(\hat y)
         +
         d(x, \hat x_\beta) 
         + \\ 
         & \quad %
         \beta \big(
          \underbrace{%
            -\log(D(\hat y, \hat x_\beta)%
          }_{\mathcal L_G}
         +
         C_P \mathcal L_P (x, \hat x_\beta)
         \big)
    \Big].
\end{split}
\end{equation}
We use $d(x, \hat x_\beta){=}1/100\,\text{MSE}(x, \hat x_\beta)$ (where MSE is calculated on inputs and reconstructions scaled to $\{0, \dots, 255\}$), and $\lambda '{=}100 \lambda$.
Following~\cite{mentzer2020high}, we use $\mathcal L_P{=}\text{LPIPS}$~\cite{zhang2018unreasonable}.
$C_P$ is a hyper-parameter to weight $\mathcal L_P$ relative to $\mathcal L_G$.
We note that in contrast to Eq.~\ref{eq:lossmse}, we have the rate parameter $\lambda'$ on $r(\hat y)$ instead of MSE. This formulation allows us to target different bitrates without changing the relative weight of the distortion compared to other terms. The factor $1/100$ in $d$ and $\lambda '$ is chosen such that this formulation is the same as Eq.~\ref{eq:lossmse} for $\lambda = 1/100$.

During training, we uniformly sample $\beta$ and minimize 
$\mathbb E_{\beta\sim U(0, \beta_\text{max})}\mathcal L_{EGD}(\beta)$, using $\beta_\text{max}{=}5.12$ for training.
During inference the receiver can choose $\beta$ freely to navigate the distortion-realism trade-off, obtaining different $\hat x_\beta$ from a fixed $\hat y$. 
As motivated in Sec.~\ref{sec:ablations}, we use $\beta_\text{max}{=}2.56$ for inference.

\subsection{Beta Conditioning}\label{sec:betacond}

To condition $G$, we use the $\beta$-conditioning scheme shown in Fig.~\ref{fig:archoverview}, which we call \textbf{FourierCond}.
It is inspired by how diffusion models condition on the timestep\cite{ho2020denoising}:
We first obtain global (\ie shared for all layers) features $f(\beta)$ by calculating Fourier features~\cite{vaswani2017attention, mildenhall2021nerf}.
Here, we use the NeRF~\cite{mildenhall2021nerf} approach, using $L{=}10$ in~\cite[Eq.~2]{mildenhall2021nerf}.
Afterwards, we apply a 2-layer MLP (with ReLU activations and 512 channels for each of the two dense layers).
We then learn a projection for each convolutional layer in each residual block in our $G$.

To explore whether it matters how exactly $\beta$ is fed into $G$, we explore a second approach in the ablations, which we named \textbf{TableCond}. It is inspired by multi-rate image compression models~\cite{balle2020nonlinear}, where we use a lookup table indexed by $\beta$ to obtain scaling factors and biases which are applied to each of the (non-residual) convolutions in $G$.

We note that since we restrict our inference to a finite set of $\beta$s, both conditioning schemes lead to the same runtime during inference.
After all, for any $\beta$, we can pre-compute the offset that gets applied to the residual blocks.

\section{Experiments}
\label{sec:experiments}

\subsection{Metrics} 
We evaluate our models for distortion and realism through PSNR and FID~\cite{heusel2017gans} respectively.
PSNR in RGB is still the most widely used metric to asses distortion-optimized neural codecs, whereas FID is widely used to asses generative models in terms of realism~\cite[\dots]{ramesh2021zero,yu2022scaling,saharia2022photorealistic}.

\subsection{Datasets} 
We train our method on 256px crops extracted from a large set of high-resolution images, where each image is randomly resized such that the shorter side is between 500 and 1000 pixels.
We evaluate on the following common benchmark datasets from \emph{image compression}: \textbf{Kodak}~\cite{kodakurl}, 24 images of resolution $512{\times}768$ or $768{\times}512$ and \textbf{CLIC 2020}~\cite{clic2020}, from which we use the \texttt{test} split with
428 high-resolution images. The shorter side is $\approx$1500px for most images (see~\cite[Fig.~A12]{mentzer2020high} for more statistics).
For Kodak, we only report PSNR since it has too few images to reliably estimate FID.
For CLIC 2020, we follow HiFiC~\cite[Sec.~A.7]{mentzer2020high} and report \emph{patched} FID, where we extract $256{\times}256$ patches that cover all images (which we denote ``FID/256''). 
This produces ~30K overlapping patches (and ~15K unique patches), which is of the order of magnitude required to measure FID. 
However, neither of these datasets are commonly used for evaluation in the generative modeling literature (\eg, DALL-E~\cite{ramesh2021zero}, Parti~\cite{yu2022scaling} and Imagen~\cite{saharia2022photorealistic}),
where \textbf{MS-COCO 30K} has become the main benchmark dataset, which we thus also use.
The dataset consists of 30\,000 $256{\times}256$ images obtained from the MS-COCO 2014 validation 
set.\footnote{%
Like Parti~\cite{yu2022scaling}, we use the Dalle processing~\cite[Sec.~A.2,~Listing~1]{ramesh2021zero}.}
We note that FID/256 is equivalent to vanilla FID on COCO.

\subsection{Building Strong  Baselines}\label{sec:strongbase}

No code or set of reconstructions is publicly available for the state-of-the-art non-generative image compression methods, so we aim to match the approach of He~\etal, ELIC~\cite{he2022elic} in PSNR,
as it is state-of-the-art while reporting fast inference on GPU (50ms on a Nvidia Titan XP for a $512{\times}768$ image).\footnote{%
Koyuncu~\etal~\cite{koyuncu2022contextformer} report marginally better PSNR at a significant increase in inference time.}
We use their $E, G$, but like for our method (Sec.~\ref{sec:ourapproach}), we use $N{=}256$ for $G$,
and also use the Charm~\cite{minnen2020channel} entropy model (\ie, in contrast to the paper by He~\etal, we use equally sized slices and no checkerboard). 
The resulting model almost matches ELIC in PSNR (there is a ${\approx}0.1dB$ difference on Kodak, see Sec.~\ref{sec:results}), and we thus use it as a stand-in for state-of-the-art in PSNR, calling it \textbf{SOTA MSE Baseline}.

We combine this model with the discriminator and GAN formulation from Sec.~\ref{sec:ourapproach} to form our \textbf{GAN baseline}. We train it for Eq.~\ref{eq:lossbeta} using a fixed $\beta{=}2.56$, \ie, this can be viewed as the same as our main model but using a non-conditional $G$ that can only target a single realism weight.
We use this baseline to tune the weights $C_P$ for LPIPS and the GAN weight $\beta$, reported in Sec.~\ref{sec:ablations}, and then use the resulting $C_P$ for our main models.
The GAN baseline method outperforms HiFiC on COCO in FID, and nearly matches it on CLIC 2020, while being significantly stronger in terms of PSNR.
We note that this is despite the fact that we do not port the multi-stage training or rate controller from HiFiC, \ie, we train our models end-to-end from scratch.

Theoretically, a stochastic decoder is necessary to achieve high perceptual quality at low rates~\cite{tschannen2018deep}, so we explored concatenating noise to the representation before decoding in the GAN baseline.
However, we found that this did not affect reconstructions at the rates we are interested in 
(intuitively stochasticity is crucial as the bitrate approaches zero).

\subsection{Published Baselines}

Since the publication of HiFiC~\cite{mentzer2020high} in 2020, there has been limited research in improving generative image compression. The few methods that have been published~\cite{po-elic,Ma2022a,iwai2021fidelity} have very limited evaluation (usually only on the validation sets of CLIC'21 (41 images) or '22 (30 images) which do not have enough images to estimate FID), and do not publish code to run on custom datasets.
In contrast, \textbf{HiFiC} has code and reconstructions available.
On high-resolution datasets large enough to estimate FID (CLIC 2020 and MS-COCO 30K), HiFiC remains state-of-the-art in terms of FID prior to the presented work.

On the MSE side, we compare to Minnen~\etal's~\textbf{Charm}~\cite{minnen2020channel}, since code is available for it and it is still close to state-of-the-art.
We compare to \textbf{ELIC}~\cite{he2022elic} in terms of PSNR on Kodak, as well as visually to the two reconstructions they publish in Sec.~\ref{sec:app:eliccmp}.
We additionally compare to
the non-learned \textbf{BPG}~\cite{bpgurl} (based on the HEVC standard) and
\textbf{VTM}~17.1~\cite{vtm17} (the reference implementation of VVC~\cite{bross2021overview}).
VVC/VTM is the state-of-the-art among non-neural image codecs. 
In Sec.~\ref{sec:app:iwaicmp} we zoom into the lowest bitrates to compare to \textbf{Iwai} et al. \cite{iwai2021fidelity}.

We detail how we run the publicly available methods in Sec.~\ref{sec:app:codecinfo}.

\subsection{Our Models}
We train all baselines and ablations for 2M iterations at batch size $8$ on 256px crops.
For the multi-realism models, we train for 3M steps since the decoder needs to simultaneously learn to achieve high and low realism (we note that 3M is still less than training two models that target a single $\beta$, and our model can target infinitely many $\beta$s.)
We use the Adam optimizer, with learning rate $1\sc{e}^{-4}$ and default settings.
As common in the literature, we train with a higher lambda (10x) in the first 15\% steps, and decay the learning rate by a factor 10x in the last 15\% steps. We did not tune these training parameters.
We evaluate our model for $\beta \in \{0.0, ..., 2.56\}$ (motivated in Sec.~\ref{sec:ablations}).

\section{Results}
\subsection{Main Results}

\begin{figure}
    \centering
    \includegraphics[width=\linewidth]{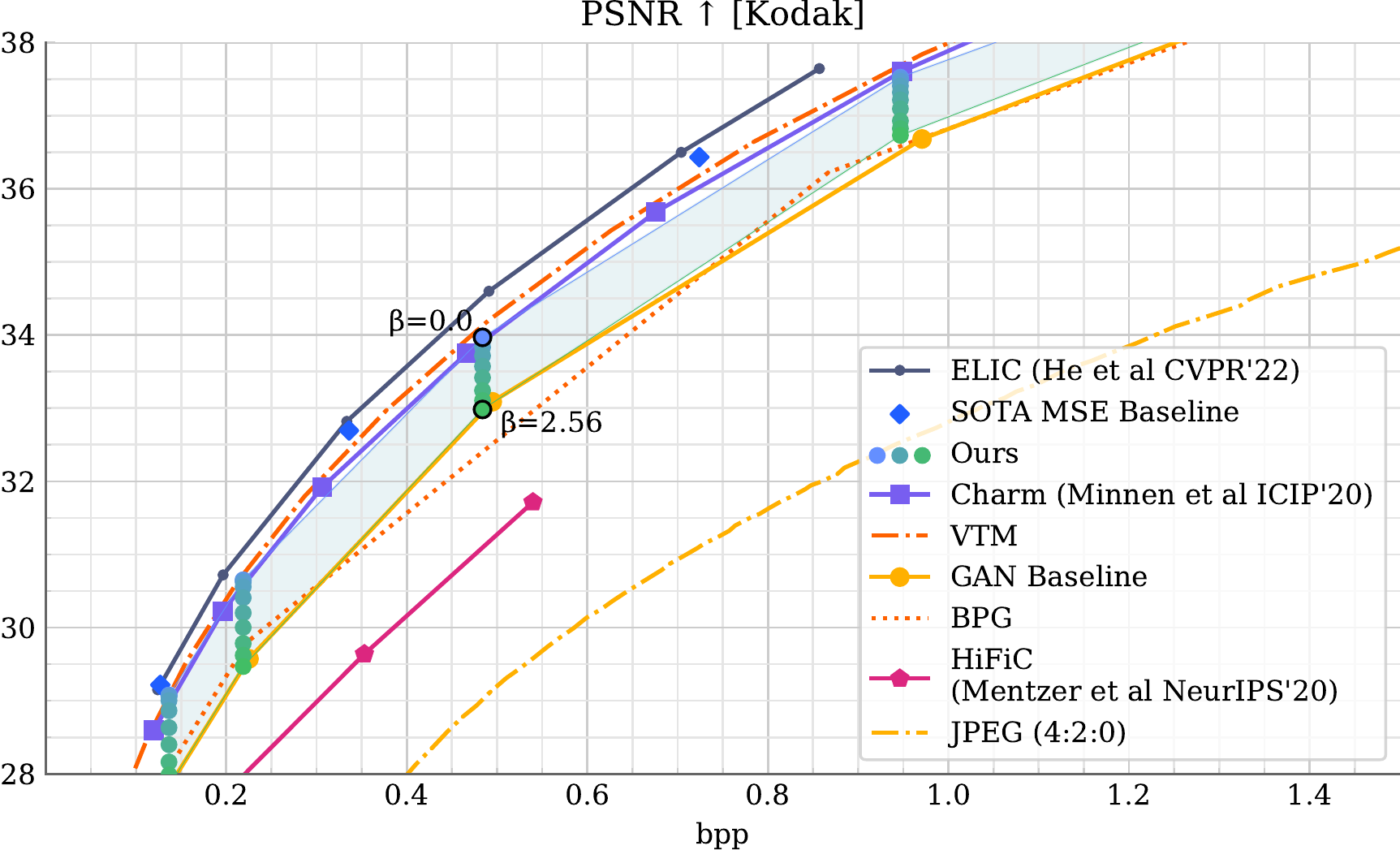}
    \caption{Results on Kodak in terms of PSNR (note that Kodak is too small (24 images) to estimate FID).
    Our multi-realism model is strong in terms of PSNR in the low-$\beta$ mode while also allowing for high realism reconstructions (see Fig~\ref{fig:visualcomp2}). 
    Additionaly, here we can see that our ``SOTA MSE Basline'' is competitive with the published (state-of-the-art) ELIC\cite{he2022elic} model, we observe only a tiny gap ($0.0-0.2$dB).
    \label{fig:resultskodak}
    }
\end{figure}

As mentioned in Sec.~\ref{sec:strongbase}, the state-of-the-art in image compression~\cite{he2022elic,koyuncu2022contextformer} in terms of MSE 
does not provide code or reconstructions.
We thus use our ``SOTA MSE Baseline'' as a stand-in for the state-of-the-art in terms of PNSR.
We establish its strength in Fig.~\ref{fig:resultskodak}, where we show that it is ${\approx}0.0-0.2$dB below ELIC~\cite{he2022elic}.

In Fig.~\ref{fig:results}, we show that our model can achieve a new state-of-the-art in terms of distortion-realism:
On the high-realism side ($\beta{=}2.56$), we 
match or outperform the state-of-the-art generative method HiFiC in FID (left plots, note the annotation of $\beta$), while also
significantly outperforming it in PSNR (right plots).
On the low-distortion side ($\beta{=}0$), 
we are strong in PSNR, reaching towards the SOTA MSE baseline in terms of PSNR (right plots), while significantly outperforming it in FID (left plots).
We emphasize that this means that 
\textit{a)} our model is significantly closer to the input than HiFiC in the high-$\beta$ mode (\ie, it has higher PSNR, see right plots), leading to more faithful reconstructions,
and also 
\textit{b)} we have greater realism than state-of-the-art MSE models in the low-$\beta$ mode.

This is even more apparent as we consider Fig.~\ref{fig:dist_perc}, which shows a single rate point and compares PSNR vs.\ FID.
In this figure, it is best to be in the upper left, like in the common rate-distortion plots. 
\emph{We can see that our approach reaches closer to this optimum than any previous method}.
We now see more clearly that for low FID (high $\beta$), our  method has significantly higher PSNR than HiFiC ($\approx 1dB$), while for high PSNR (low $\beta$), our method has signficantly lower FID (about $40\%$) than any of the non-generative methods (VTM, BPG, Charm, and the SOTA MSE baseline).

Comparing our model at $\beta=0$ to the MSE baseline and our model at $\beta=2.56$ to the GAN basline, we might expect symmetric gaps on both sides. However, as we can see in Fig.~\ref{fig:dist_perc}, 
our GAN baseline leads to a similar operating point as our model set to $\beta{=}2.56$, 
while the MSE basline has slightly better PSNR. %
It appears that 
the multi-task nature of our loss leads to models that slightly favor realism, perhaps not surprisingly, given that we randomly sample $\beta$ during training and a large portion of the optimization thus uses $\beta{>}0$. Indeed, our model at $\beta{=}0$ actually has significantly better FID than the MSE baseline.

\begin{figure*}
    \centering
    \includegraphics[width=\linewidth]{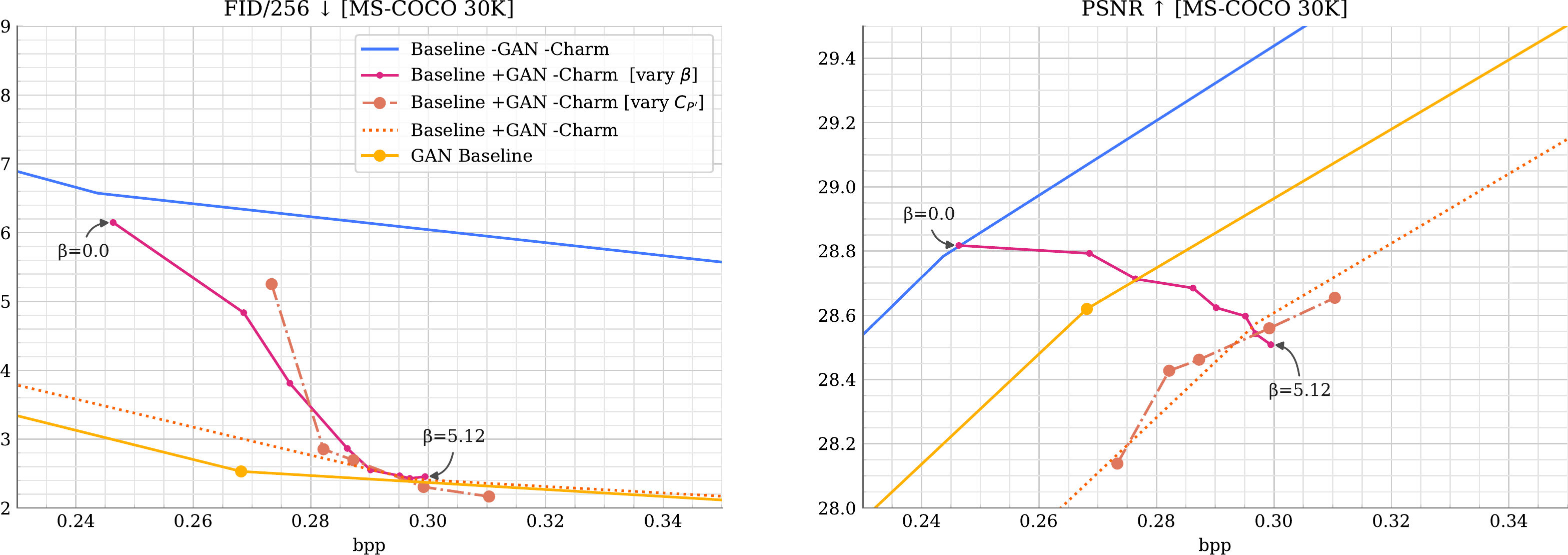}
    \caption{
    \label{fig:ablations}
    Ablations and tuning for our GAN Baseline. See Sec.~\ref{sec:ablations} for details.
    \vspace{-2ex}
    }
\end{figure*}

Additionally, in Fig.~\ref{fig:resultskodak}, we compare PSNR on Kodak,
a commonly used benchmark in compression.
Here we can see that our ``SOTA MSE Baseline'' is competitive with state-of-the-art ELIC~\cite{he2022elic}.
Furthermore, despite the fact that we train a \emph{single} model for the full spectrum from high-realism to low-distortion, our method at $\beta{=}0$ is competitive with VTM~\cite{bross2021overview} and Charm~\cite{minnen2020channel} in terms of PSNR, and only ${\approx}0.5dB$ below the state-of-the-art (with significantly better realism as discussed above).

\paragraph{Visual Comparison}

We compare visually to the generative method HiFiC, our strongest contender, in Fig.~\ref{fig:visualcomp2}.
We can observe that our reconstructions $\beta=2.56$ are closes to the input than HiFiC when we compare at the same bitrate (first two rows), or we achieve the same visual quality at lower bitrates (third and forth row). Our model achieves higher PSNR as we go towards $\beta{=}0$, but we lose detail in the reconstructions, as expected.
For completeness, we compare to the two reconstructions available for the non-generative state-of-the-art ELIC in Sec.~\ref{sec:app:eliccmp}. Visually, the reconstructions look similar to our model at $\beta{=}0$.
Finally, we provide reconstructions on CLIC 2020 in Sec.~\ref{sec:app:clicrecons}.
\subsection{Ablations \& Tuning}\label{sec:ablations}

In Fig.~\ref{fig:ablations} we show the results for our ablation experiments on MS-COCO.
For the ablations, we decouple the weight on $\mathcal{L}_P$ in Eq.~\ref{eq:lossbeta} by setting $\beta C_P = C_P'$.

\subsubsection{Loss Weights and Charm} 
Most ablation experiments are conducted using the Hyperprior~\cite{minnen2018joint} entropy model instead of Charm~\cite{minnen2020channel}, which we denote with ``-Charm''.
We trained the baseline for various rate points with a fixed LPIPS weight $C_P' = 4.26$  (the default weight of HiFiC\cite{mentzer2020high} ) but without the GAN loss, which we refer to as ``Baseline -GAN -Charm''.
We then introduce the GAN loss, 
and vary first the weight of the GAN loss $\beta \in \{0.0, 0.08, 0.16, 0.32, \cdots 5.12\}$
(``Baseline +GAN -Charm [vary $\beta$]'').
Here we find that the GAN loss lowers the FID up to $\beta=2.56$ where it starts saturating.
With $\beta$ fixed to this value, we now vary $C_P'\in \{0.0, 1.0, 2.0, 4.26, 8.0\}$ and find that the weight $4.26$ remains a good choice (``Baseline +GAN -Charm [vary $C_{P'}$]''). 
The resulting model is called ``Baseline +GAN -Charm''.
Now we adopt Charm\cite{minnen2020channel} entropy modeling 
which lowers the bitrate (resulting in ``GAN Baseline'').

For the main model, we sampled $\beta$ uniformly in the range $[0, 5.12]$ during training,
but also found that the FID score at inference is lowest for $\beta{=}2.56$. 
We thus set $\beta_\text{max}{=}2.56$ for our model during inference. 

\subsubsection{Conditioning} When comparing \textit{FourierCond} and \textit{TableCond} we found they give very similar results, see Sec.~\ref{sec:app:table_vs_mlp}.
We choose \textit{FourierCond} for our main model as it leads to a simpler implementation.

\label{sec:results}

\vfill

\section{Conclusion}

We have presented a method which is capable of outputting a single representation for compressed images, from which a receiver can either decode a high-realism reconstruction (high $\beta$) or a high-PSNR reconstruction (low $\beta$). We saw that in terms of distortion (PSNR) vs.\ realism (FID), our method can reach a new state-of-the-art. 
To the best of our knowledge, this is the first single decoder method which allows the trade-off between
realism and distortion to happen on the receiver side, with no change in the bitstream.
This means that depending on the use-case, the users may choose to view reconstructions which are as close to the original as possible, or switch to view images with a better level of (generated) detail.

Somewhat surprisingly, we find that we can obtain high realism without sacrificing PSNR by more than ${\approx}1-1.5$dB.
We hope our findings inspires further work to push the boundary of the the realism-distortion trade-off.

\FloatBarrier
\newpage

{\small
\bibliographystyle{ieee_fullname}
\bibliography{egbib}
}

\clearpage
\newpage

\appendix

\section{Supplementary Material -- Multi-Realism Image Compression with a Conditional Generator}

\input{runtime}

\subsection{Running Published methods} \label{sec:app:codecinfo}

\paragraph{HiFiC} We use the models published in Tensorflow Compression~\cite{tfc_github}, using \texttt{tfci} to run the models called \texttt{hific-lo}, \texttt{hific-mi}, \texttt{hific-hi}

\paragraph{VTM} We use VTM 17.1~\cite{vtm17}.
To compress images, we convert them to YCbCr, store the result as raw bytes in at \texttt{\$YUVPATH} and then run the following:

\noindent\begin{minipage}{\linewidth}
\vspace{1.2ex}
\begin{verbatim}
# Encode
EncoderApp -c encoder_intra_vtm.cfg 
 -i $YUVPATH -q $Q, -o /dev/null 
 -b $OUTPUT
 --SourceWidth=$WIDTH 
 --SourceHeight=$HEIGHT
 --FrameRate=1 --FramesToBeEncoded=1
 --InputBitDepth=8 
 --InputChromaFormat=444
 --ConformanceWindowMode=1
 
# Decode
DecoderApp 
 -b $OUTPUT -o $RECON -d 8
\end{verbatim}
\end{minipage}

\paragraph{BPG} We use BPG v0.9.8 and run the following given input PNGs:

\noindent\begin{minipage}{\linewidth}
\vspace{1.2ex}
\begin{verbatim}
# Encode
bpgenc -o $OUTPUT -q $Q
  -f 444 -e x265 -b 8 $SRC
  
# Decode
bpgdec -o $RECON $OUTPUT
\end{verbatim}
\end{minipage}

\subsection{Comparison to Iwai et al.}\label{sec:app:iwaicmp}
\begin{figure*}
    \centering
    \includegraphics[width=0.6\linewidth]{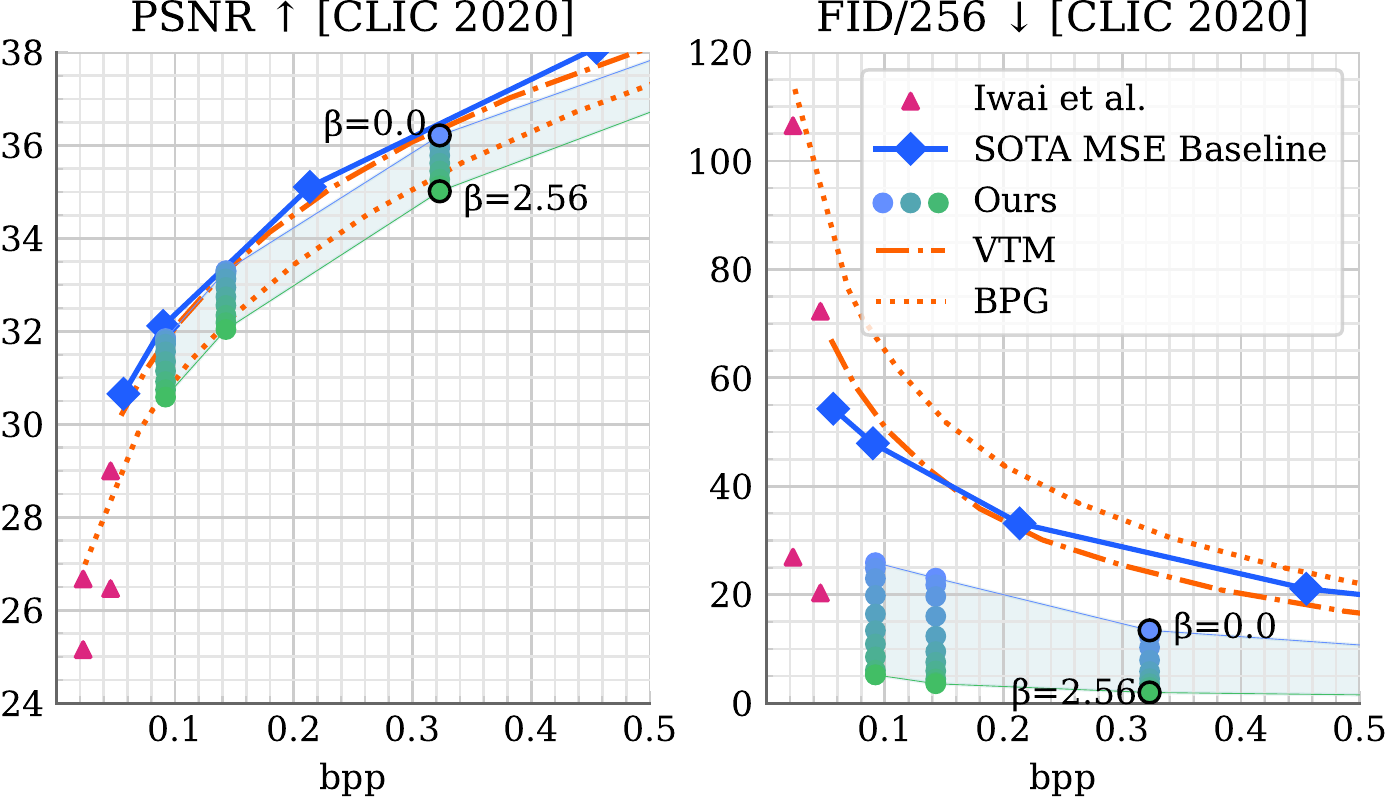}
    \caption{\label{fig:iwai_comparison}
    Comparison to Iwai et al.\cite{iwai2021fidelity}.
    We can see that they target lower bitrates but this results in significantly worse PSNR and FID.
    }
\end{figure*}

To compare with \cite{iwai2021fidelity}, we ran their released model on CLIC and evaluated their model in terms of FID and PSNR versus bitrate, see Figure~\ref{fig:iwai_comparison}.

\subsection{Comparing to ELIC Reconstructions} \label{sec:app:eliccmp}

\begin{figure*}
\begin{tabular}{@{\hskip 0mm}l@{\hskip 0.5mm}l@{\hskip 1.5mm}l@{\hskip 0.5mm}l@{\hskip 0mm}}

\multicolumn{2}{c}{High-Realism} &
\multicolumn{2}{c}{Low-Distortion} %
\\\midrule

    Ours $\beta{=}2.56$ &
    HiFiC \cite{mentzer2020high} &
    Ours $\beta{=}0.0$ &
    ELIC \cite{he2022elic} \\
    \includegraphics[width=0.245\linewidth]{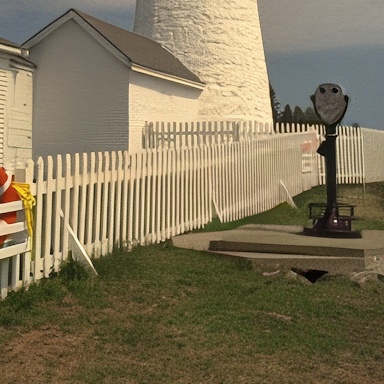} &
    \includegraphics[width=0.245\linewidth]{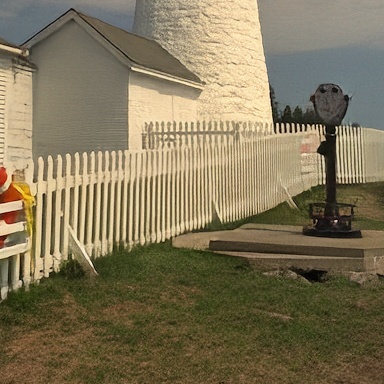}  &
    \includegraphics[width=0.245\linewidth]{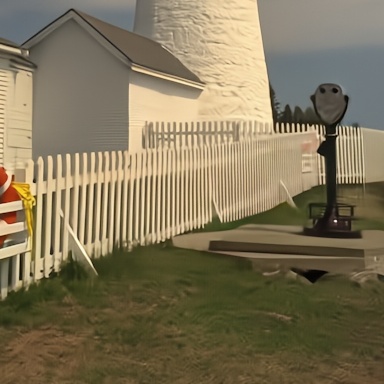} &
    \includegraphics[width=0.245\linewidth]{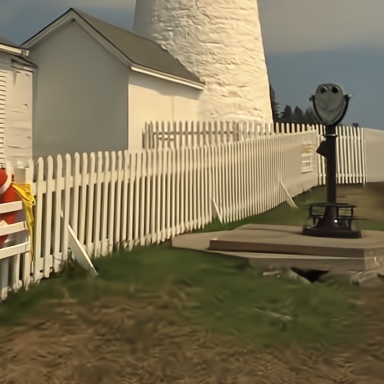} \\[-3ex]
   {\color{white}\footnotesize \hphantom{i}0.11bpp, 28.1dB} &
   {\color{white}\footnotesize \hphantom{i}0.16bpp{\color{white}($\mathbf{1.4{\boldsymbol\times}}$)}, 27.8dB} &
   {\color{white}\footnotesize \hphantom{i}0.11bpp, 29.3dB} &
   {\color{white}\footnotesize \hphantom{i}0.14bpp{\color{white}($\mathbf{1.3{\boldsymbol\times}}$)}, 30.4dB} \\
    
\includegraphics[width=0.245\linewidth]{%
  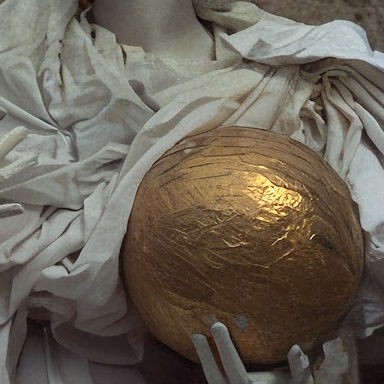} &
\includegraphics[width=0.245\linewidth]{%
  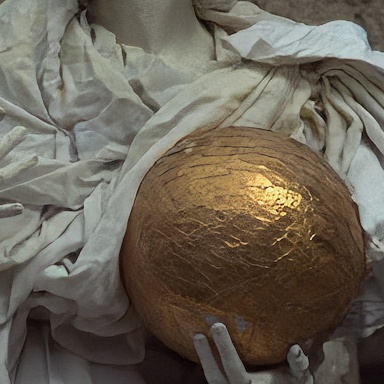} &
\includegraphics[width=0.245\linewidth]{%
  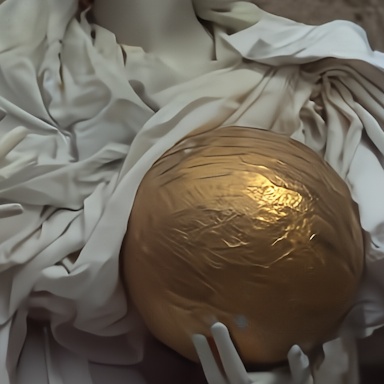} &
\includegraphics[width=0.245\linewidth]{%
  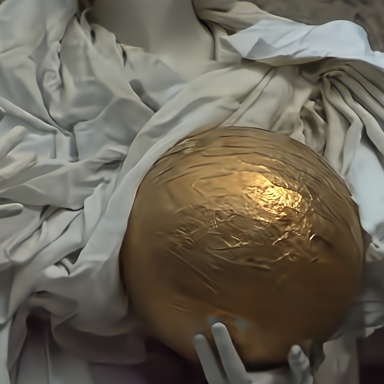}\\[-3ex] 
   {\color{white}\footnotesize \hphantom{i}0.10bpp, 29.1dB} &
   {\color{white}\footnotesize \hphantom{i}0.16bpp {\color{white}($\mathbf{1.6{\boldsymbol\times}}$)}, 29.1dB} &
   {\color{white}\footnotesize \hphantom{i}0.10bpp, 30.2dB} &
   {\color{white}\footnotesize \hphantom{i}0.13bpp{\color{white}($\mathbf{1.3{\boldsymbol\times}}$)}, 31.7dB} \\[-2ex]
\end{tabular}
\caption{\label{fig:elicvisuals}
Comparing to the published
HiFiC~\cite{mentzer2020high} and ELIC~\cite{he2022elic},
both of which are state-of-the-art for realism and distortion, respectively.
We show two reconstructions from our method, both obtained from the \emph{same} representation, where we only vary the realism weight used by the generator.
Note that PSNR increases as we move from a generative method to a purely MSE-optimized method, but our approach shows a higher PSNR than HiFiC even when using a high realism weight ($\beta=2.56$) that leads to comparable perceptual quality. \textbf{This figure is best viewed zoomed in.}
}
\end{figure*}

We compare to the two published ELIC reconstructions in Fig.~\ref{fig:elicvisuals}.

\subsection{More Realism rates} \label{sec:app:dist_realism_rates}

\begin{figure*}
    \centering
    \includegraphics[width=0.49\linewidth]{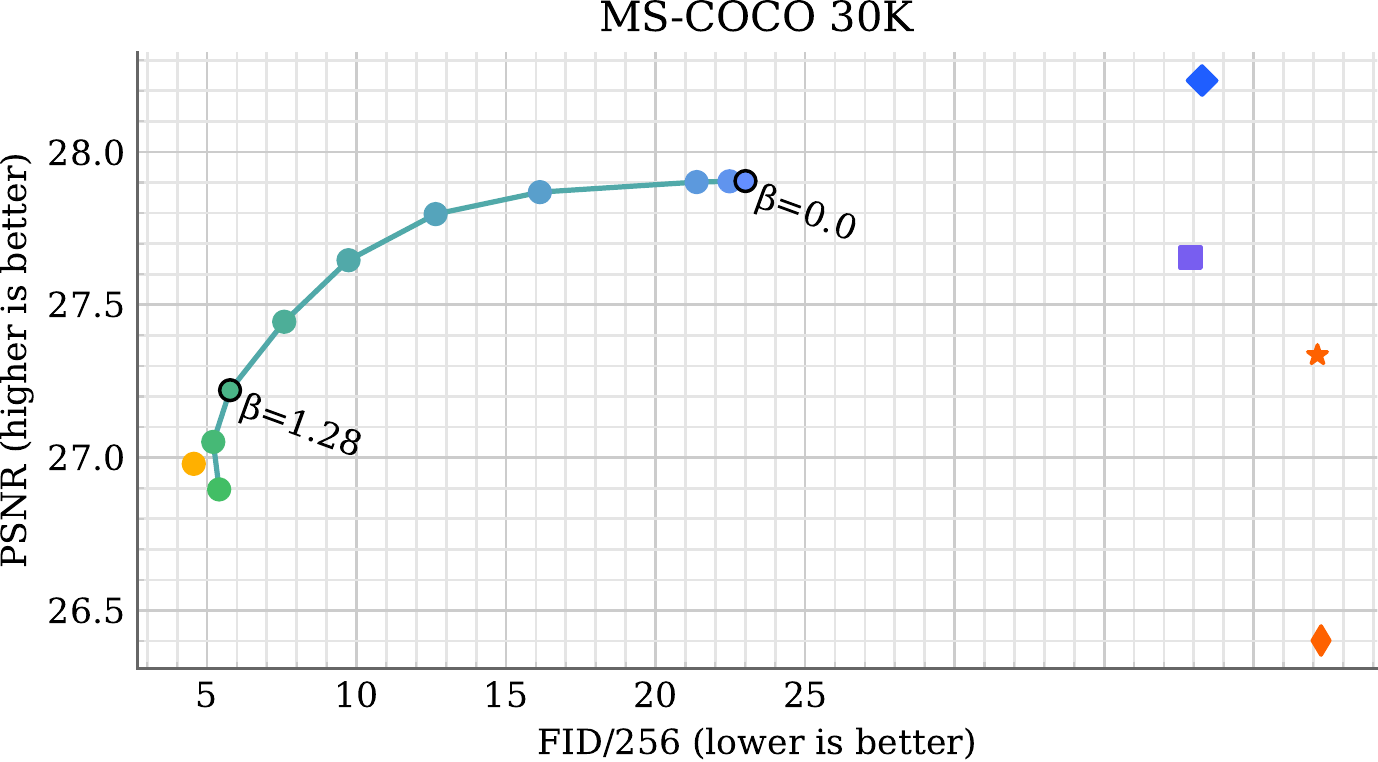}
    \includegraphics[width=0.49\linewidth]{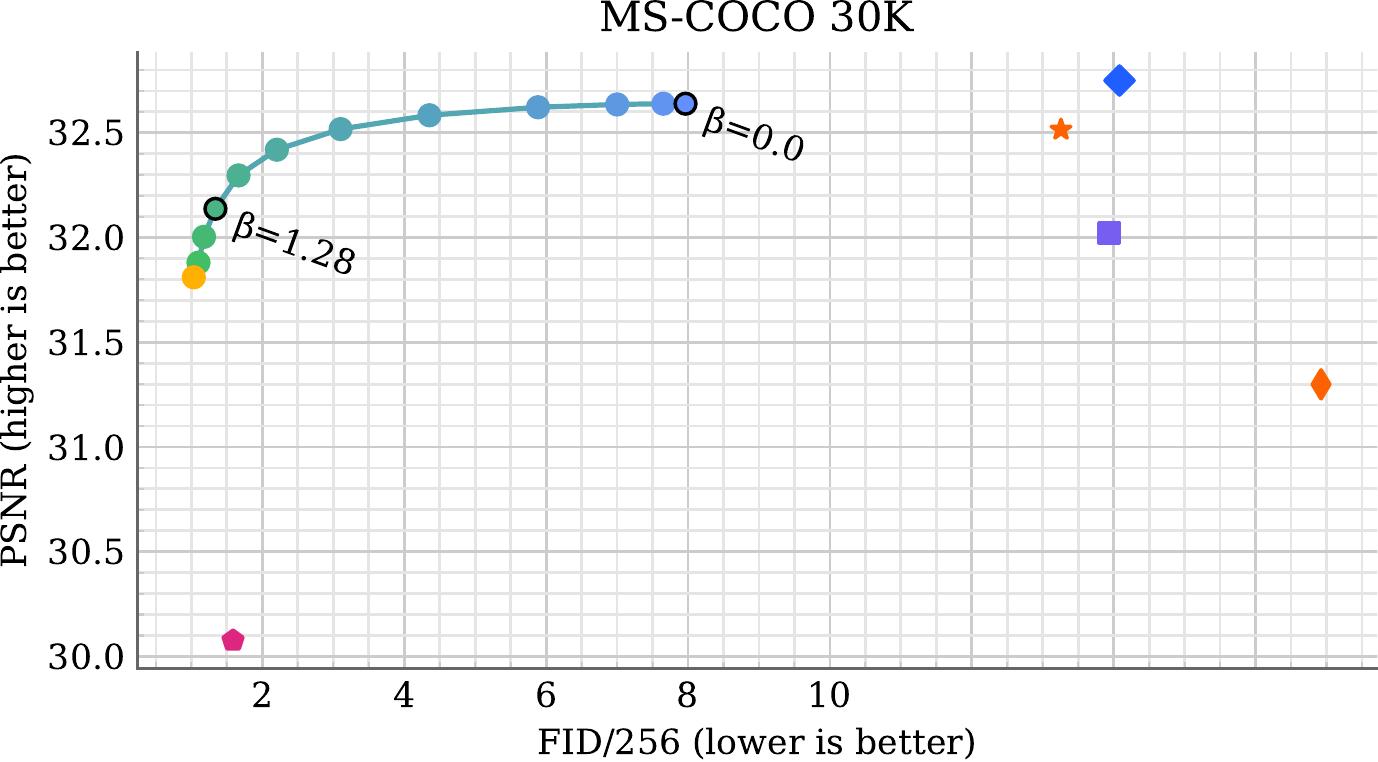} \\
    \includegraphics[width=0.49\linewidth]{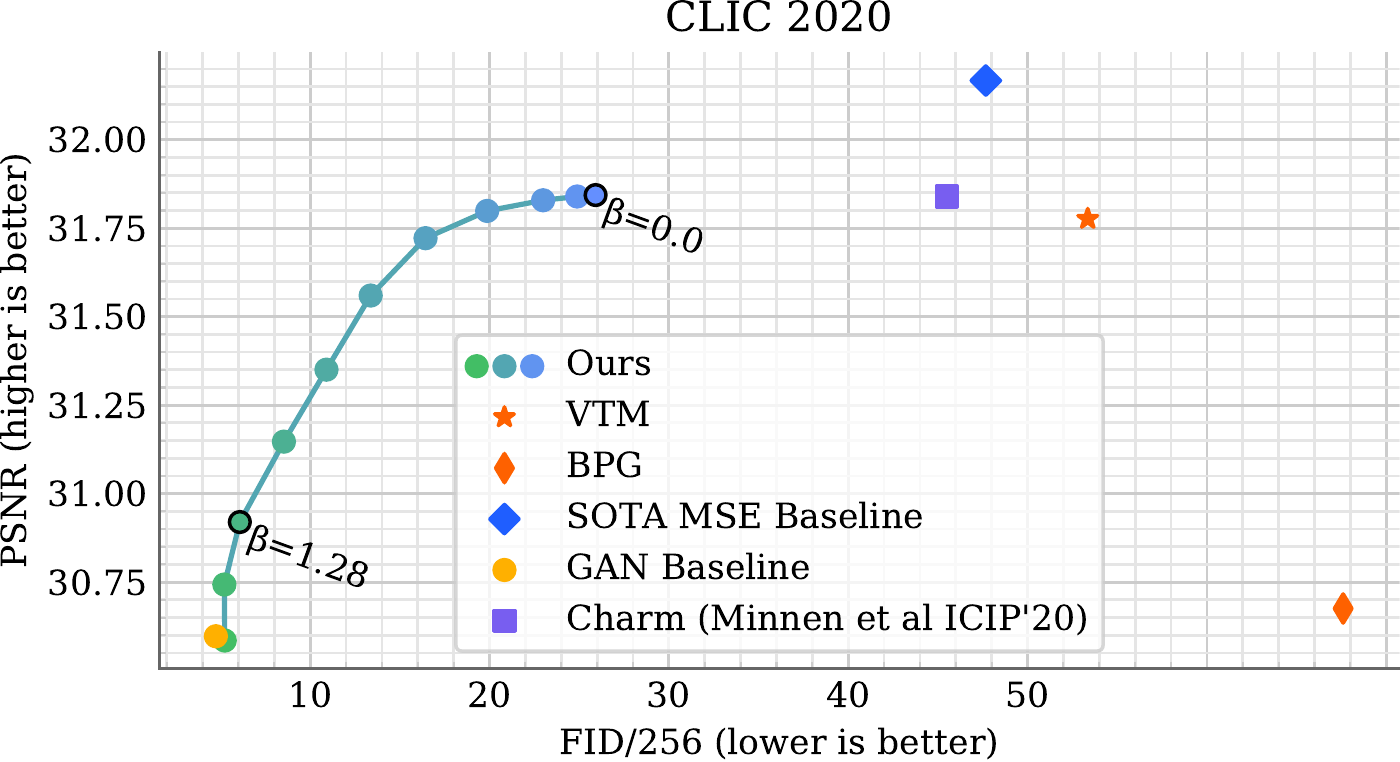}
    \includegraphics[width=0.49\linewidth]{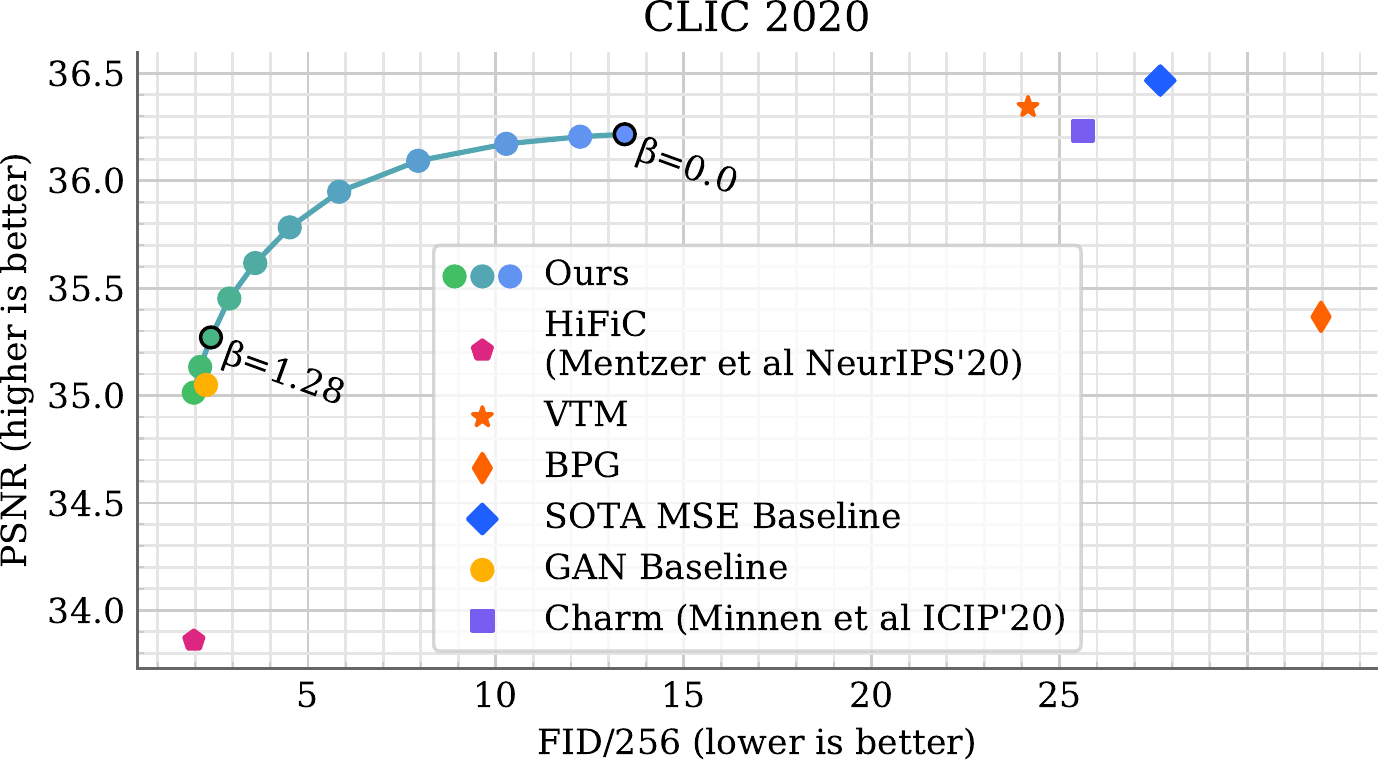}
    \caption{\label{fig:distpercmore}
    More rate points for distortion-realism, extending Fig.~\ref{fig:dist_perc}.
    The left column is at 0.17bpp for MS-COCO 30K, and 0.092bpp for CLIC 2020, whereas the right column is at 0.56bpp for MS-COCO 30K and 0.32bpp for CLIC 2020.
    }
\end{figure*}

We show Fig.~\ref{fig:dist_perc} at more rate poitns in Fig.~\ref{fig:distpercmore}.

\subsection{Comparing conditioning schemes} \label{sec:app:table_vs_mlp}
\begin{figure*}
    \centering
    \includegraphics[width=\linewidth]{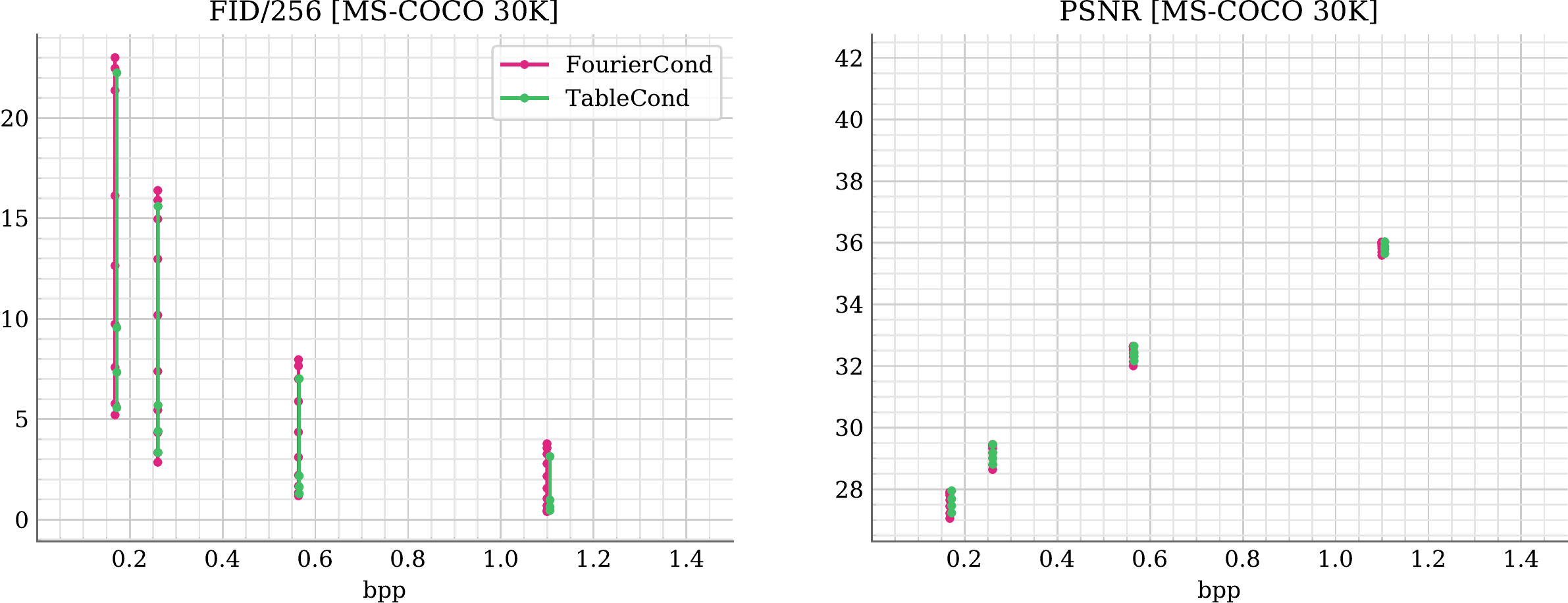}
    \caption{
    \label{fig:table_vs_mlp}
    We compare the FourierCond and TableCond conditioning approaches for $G$. We find both give comparable results and choose FourierCond for our main model.}
\end{figure*}

In Figure~\ref{fig:table_vs_mlp} we compare the FourierCond and TableCond conditioning approaches. Both methods were trained in the same manner as our main model.

\subsection{Visual Results}
\label{sec:app:clicrecons}

We provide CLIC 2020 reconstructions for our lowest rate model, both for $\beta=0.0$ and $\beta=2.56$ in a zip file which have uploaded here:
\href{https://storage.googleapis.com/multi-realism-paper/multi\_realism\_paper\_supplement.zip}{https://storage.googleapis.com/multi-realism-paper/multi\_realism\_paper\_supplement.zip}

Additionally, we  include the MD5 checksum of the file to allow for verifying its integrity:
{
\begin{verbatim}
$ md5sum \
 multi_realism_paper_supplement.zip

2ed64ff793bbb87b7a884750ab61c912  
 multi_realism_paper_supplement.zip
\end{verbatim}
}

\end{document}

%% file: runtime.tex
\subsection{Runtime Benchmarks}
We present runtime benchmark results on three systems. The first system used an NVIDIA Tesla V100, released in late 2017. The results for this system are presented in Table~\ref{tab:runtime_v100}. The second system employs an NVIDIA 2080 Ti, a consumer GPU targeted primarily at video gaming and released 2018. The results for this system are summarized in Table~\ref{tab:runtime_2080}.
The third system uses a more powerful NVIDIA 3090 Ti GPU, summarized in Table~\ref{tab:runtime_3090ti}.

To present a realistic usage of the method, we include the CPU runtime needed to encode images (i.e., range coding is included in the total compression/entropy encoding/decoding/total decompression time).

We note that comparing runtime numbers across papers is challenging due to implementation details and platform-specific optimizations.
In our case, we did not aim to provide the fastest runtime numbers and, as such, when implementing the ELIC method~\cite{he2022elic}, we omitted some performance critical features. Primarily, we did not use checkerboard encoding, nor did we employ uneven grouping for CHARM. Both of these should provide better runtime performance for the entropy coding and decoding components.

\begin{table*}[t]
\centering
\small
\begin{tabular}{lcc|cc|cccc}
\toprule
Model  &  Encoder &  \adjustbox{stack=cc}{Entropy\\Coding} &
 \adjustbox{stack=cc}{Entropy\\Decoding} & Decoder &
 \multicolumn{2}{c}{\adjustbox{stack=cc}{Total\\Compression}} & \multicolumn{2}{c}{\adjustbox{stack=cc}{Total\\Decompression}}   \\
 & [ms] 
 & [ms] 
 & [ms] 
 & [ms] 
 & [ms] 
 & [mp/s] 
 & [ms] 
 & [mp/s] 
 \\
 \midrule
Ours (FourierCond)        &        121.3 &                   55.6 &                   50.6 &        153.8 &              176.8 &                11.2 &              204.4 &                 9.7 \\
MSE Baseline (N=192)      &         71.2 &                   52.3 &                   47.6 &         86.3 &              123.6 &                16.0 &              133.9 &                14.8 \\
SOTA MSE Baseline (N=256) &        121.4 &                   66.6 &                   74.3 &        139.0 &              188.0 &                10.5 &              213.3 &                 9.3 \\
HiFiC                     &         43.2 &                   47.4 &                   57.4 &        200.5 &               90.6 &                21.9 &              257.9 &                 7.7 \\

\bottomrule
\end{tabular}

\caption{Runtime numbers (ms) needed to process a one megapixel image on a \textbf{Tesla V100} using float32 precision. Compression/decompression numbers include predicting entropy parameters on the GPU and running the range coder/decoder on CPU.}
\label{tab:runtime_v100}
\end{table*}




\begin{table*}[t]
\centering
\small
\begin{tabular}{lrr|rr|rrrr}
\toprule
Model  &  Encoder &  \adjustbox{stack=cc}{Entropy\\Coding} &
 \adjustbox{stack=cc}{Entropy\\Decoding} & Decoder &
 \multicolumn{2}{c}{\adjustbox{stack=cc}{Total\\Compression}} & \multicolumn{2}{c}{\adjustbox{stack=cc}{Total\\Decompression}}   \\
 & [ms] 
 & [ms] 
 & [ms] 
 & [ms] 
 & [ms] 
 & [mp/s] 
 & [ms] 
 & [mp/s] 
 \\
 \midrule
Ours (FourierCond)        &        128.4 &                   92.3 &                   75.2 &        205.9 &              220.7 &                 9.0 &              281.1 &                 7.1 \\
MSE Baseline (N=192)      &         81.4 &                   89.1 &                   72.4 &        109.2 &              170.5 &                11.6 &              181.6 &                10.9 \\
SOTA MSE Baseline (N=256) &        128.7 &                   68.5 &                   72.9 &        155.5 &              197.2 &                10.1 &              228.4 &                 8.7 \\
HiFiC                     &         40.5 &                   48.9 &                   57.5 &        248.3 &               89.4 &                22.2 &              305.8 &                 6.5 \\
\bottomrule
\end{tabular}

\caption{Runtime numbers (ms) needed to process a one megapixel image on a \textbf{NVIDIA 2080 Ti} consumer GPU using float32 precision. Compression/decompression numbers include predicting entropy parameters on the GPU and running the range coder/decoder on CPU.}
\label{tab:runtime_2080}
\end{table*}


\begin{table*}[t]
\centering
\small
\begin{tabular}{lrr|rr|rrrr}
\toprule
Model  &  Encoder &  \adjustbox{stack=cc}{Entropy\\Coding} &
 \adjustbox{stack=cc}{Entropy\\Decoding} & Decoder &
 \multicolumn{2}{c}{\adjustbox{stack=cc}{Total\\Compression}} & \multicolumn{2}{c}{\adjustbox{stack=cc}{Total\\Decompression}}   \\
 & [ms] 
 & [ms] 
 & [ms] 
 & [ms] 
 & [ms] 
 & [mp/s] 
 & [ms] 
 & [mp/s] 
 \\
 \midrule
Ours (FourierCond)        &         62.7 &                   45.9 &                   37.2 &         80.7 &              108.6 &                18.3 &              117.9 &                16.8 \\
MSE Baseline (N=192)      &         44.6 &                   46.0 &                   37.5 &         47.9 &               90.6 &                21.9 &               85.4 &                23.2 \\
SOTA MSE Baseline (N=256) &         62.9 &                   47.2 &                   54.0 &         68.8 &              110.1 &                18.0 &              122.8 &                16.1 \\
HiFiC                     &         20.7 &                   33.2 &                   46.8 &        109.7 &               54.0 &                36.7 &              156.5 &                12.7 \\
\bottomrule
\end{tabular}

\caption{Runtime numbers (ms) needed to process a one megapixel image on a \textbf{NVIDIA 3090 Ti} consumer GPU using float32 precision. Compression/decompression numbers include predicting entropy parameters on the GPU and running the range coder/decoder on CPU. Please note that certain components don't scale linearly when compared to the \textbf{NVIDIA 2080 Ti}. This is because
entropy encoding/decoding is still serial and the autoregressive components aren't parallelized.}
\label{tab:runtime_3090ti}
\end{table*}